\providecommand\acmformat{manuscript}
\documentclass[\acmformat]{acmart}
\usepackage{array}

\pdfinterwordspaceon

\setcopyright{none}
\acmJournal{AILET}
\acmYear{2026}
\copyrightyear{2026}
\acmVolume{1}
\acmNumber{1}
\acmArticle{}
\acmMonth{10}

\begin{document}

\title[Twice as Fast, Until It Isn't]{Twice as Fast, Until It Isn't: Quantized Serving of Mid-Size Open LLMs}

\author{Smit Kadvani}
\orcid{0009-0004-7615-1080}
\author{Kriti Faujdar}
\orcid{0009-0009-5938-396X}
\author{Spurthi Tallam}
\orcid{0009-0007-6386-9255}

\affiliation{%
  \institution{Independent researcher}%
  \city{}%
  \country{United States}}
\renewcommand{\shortauthors}{Kadvani, Faujdar, and Tallam}
\authorsaddresses{Smit Kadvani, Kriti Faujdar, and Spurthi Tallam, Independent researcher, United States}

\begin{abstract}
A team serving a 9B--20B open-weight language model on vLLM or SGLang must choose a quantization recipe, and needs two numbers: the speedup over BF16 and the accuracy given up.
We assemble every public measurement of these numbers that we could verify and find that the record supplies them only in pieces.
Only two checkpoints, both INT4 builds of Gemma~3 12B, have both a vLLM timing and an accuracy score, and that score covers a single MMLU subject, one that the AWQ build was calibrated on.
The batch-1 timings predate current engine releases, AMD and Intel Gaudi have no timing in this band at all, and FP4 has none against BF16.
The timings that do exist follow one rule.
Decoding a small batch is limited by memory bandwidth, so 4-bit weights roughly double speed ($2.12\times$ for Qwen2.5-14B on an A100), but the gain vanishes once a large batch makes the GPU compute-bound ($1.02\times$ for Gemma~3 12B at batch size 256), where FP8 is the recipe that can still help.
Many published accuracy deltas, including every HumanEval delta, lie within benchmark sampling noise.
We close with a seven-item reporting checklist for future measurements.
\end{abstract}



\maketitle

\section{Introduction}

Open-weight models of 9B to 20B parameters fit on one data-center GPU in BF16 and on one consumer GPU once quantized, which makes them a common choice for self-hosted serving.
A team serving one on vLLM~\cite{kwon2023vllm} or SGLang~\cite{zheng2024sglang} must pick a quantization recipe for its GPU, and needs the speedup over BF16 and the accuracy given up.
This letter asks whether the public record answers that question for Qwen2.5-14B, Qwen3-14B, Gemma~3 12B, Phi-4, Mistral NeMo, Ministral~3 14B, and gpt-oss-20b~\cite{qwen2024qwen25,qwen2025qwen3,gemmateam2025gemma3,abdin2024phi4,mistralnemo2024,ministral3card,openai2025gptoss}.
We kept a timing if its source named the engine, GPU, checkpoint, and workload, and an accuracy score if the same source reported a BF16 score from the same harness.
We searched the vLLM, SGLang, AMD ROCm, and Intel Gaudi documentation, Hugging Face model cards from the model developers and from quantization publishers (Red Hat AI, PyTorch, Tencent AngelSlim), benchmark reports, and arXiv, pairing each model name with each recipe (FP8, INT8, AWQ, GPTQ, INT4, NVFP4, MXFP4) and each engine, up to 5 October 2026.
An AI coding assistant helped collect the sources and extract the numbers; every number was then checked against its cited source on 5 October 2026.

The record answers it only in pieces, and we make three contributions.
We explain the timings that do exist with one memory-versus-compute rule (Section~\ref{sec:perf}).
We show that many published accuracy deltas are smaller than the sampling noise of their benchmarks (Section~\ref{sec:acc}).
We grade what the evidence supports and propose a reporting checklist to close the gaps (Sections~\ref{sec:supports} and~\ref{sec:checklist}).

Earlier work covers parts of this question.
Surveys cover quantization before today's serving stacks~\cite{gholami2021survey} and efficient LLM inference broadly~\cite{zhou2024survey}, and Zhao et al.\ benchmark post-training algorithms on quality alone~\cite{zhao2025ptq}.
Kurtic et al.\ measure both accuracy and vLLM speed for FP8, INT8, and INT4, but only for the Llama~3.1 family at 8B, 70B, and 405B~\cite{kurtic2025bf16}.
Shi and Ding measure eleven recipes on TensorRT-LLM for Llama-2 and CodeLlama~\cite{shi2025characterization}.
InferenceX benchmarks vLLM and SGLang on NVIDIA and AMD GPUs continuously, but lists no model in this band~\cite{inferencex2026}.

\section{Recipes and engine support}

A recipe is written W$x$A$y$: weights stored in $x$ bits, matrix multiplications run on $y$-bit activations.
Weight-only INT4 (W4A16) dequantizes 4-bit weights into a 16-bit matmul.
GPTQ (published as OPTQ) chooses the integers with a one-shot second-order reconstruction~\cite{frantar2023gptq}, AWQ rescales the channels that see large activations~\cite{lin2024awq}, and the Marlin kernel runs the dequantized matmul, reporting up to $2.8\times$ end-to-end speedup inside vLLM~\cite{frantar2024marlin}.
W8A8 also quantizes activations, so the matmul itself runs in 8 bits: LLM.int8() routes outlier features to a 16-bit path~\cite{dettmers2022llmint8}, SmoothQuant migrates them into the weights~\cite{xiao2023smoothquant}, and FP8 uses the E4M3 format~\cite{micikevicius2022fp8}.
FP8 activation scales are either fixed after calibration (static) or computed per token at run time (dynamic).
MXFP4 stores 4-bit floats in blocks of 32 that share a power-of-two scale~\cite{rouhani2023mx}; NVFP4 is NVIDIA's 4-bit float format, executed natively on Blackwell~\cite{sglangquant2026}.

Engine support depends on the GPU.
vLLM runs FP8 W8A8 on Ada, Hopper, and AMD but not on Ampere, where its Marlin kernels still serve GPTQ, AWQ, and FP8 checkpoints as weight-only~\cite{vllmquant2026}.
SGLang supports \texttt{fp8}, \texttt{awq}, and \texttt{mxfp4} on AMD Instinct MI300X and later, and keeps its Marlin kernels CUDA-only~\cite{sglangquant2026}; AMD documents FP8 and Triton AWQ for vLLM~\cite{rocmvllm2026}.

\section{Performance: one rule explains the timings}
\label{sec:perf}

\begin{table}[!tp]
\caption{Public vLLM and SGLang timings in the 9B--20B band. Each group has its own BF16 baseline, and the speedup is relative to it (higher is better; for latency, BF16 latency divided by recipe latency). Pairs a/b give two settings of the same run. ``n.r.'' means not reported. $^\dagger$Memory comes from a Hugging Face Transformers run in the same source, because vLLM and SGLang pre-allocate GPU memory; units follow each source. $^\star$Checkpoint size from the model card~\cite{openai2025gptoss}.}
\label{tab:perf}
\begin{tabular}{@{}lccc@{}}
\toprule
Recipe & Speed or latency & Speedup vs BF16 & Memory$^\dagger$ \\
\midrule
\multicolumn{4}{@{}l}{\textit{vLLM 0.6.3, one A100 80GB, Qwen2.5-14B-Instruct, batch size 1, 2048 output tokens;}} \\
\multicolumn{4}{@{}l}{\textit{tokens/s at input length 1\,/\,30{,}720}~\cite{qwen25speed}} \\
BF16 (baseline) & 46.30\,/\,37.18 & 1.00\,/\,1.00 & 28.08\,GB \\
GPTQ-Int8 (W8A16) & 70.40\,/\,49.23 & 1.52\,/\,1.32 & 16.11\,GB \\
AWQ (W4A16) & 92.66\,/\,59.01 & 2.00\,/\,1.59 & 9.79\,GB \\
GPTQ-Int4 (W4A16) & 98.02\,/\,60.91 & 2.12\,/\,1.64 & 9.94\,GB \\
\midrule
\multicolumn{4}{@{}l}{\textit{vLLM \textup{\texttt{benchmark\_latency}}, H100, Gemma~3 12B IT, 256 input and 256 output tokens;}} \\
\multicolumn{4}{@{}l}{\textit{latency in seconds at batch size 1\,/\,256}~\cite{gemma3awqcard}} \\
BF16 (baseline) & 3.73\,/\,14.07 & 1.00\,/\,1.00 & 24.51\,GB \\
TorchAO INT4 (W4A16) & 2.73\,/\,13.81 & 1.37\,/\,1.02 & 10.37\,GB \\
AWQ-INT4 (W4A16) & 2.76\,/\,13.93 & 1.35\,/\,1.01 & 12.60\,GB \\
\midrule
\multicolumn{4}{@{}l}{\textit{SGLang 0.4.6.post1, one H20 96GB, Qwen3-14B, batch size 1, 2048 output tokens;}} \\
\multicolumn{4}{@{}l}{\textit{(prompt\,+\,output tokens)/s at input length 1\,/\,30{,}720}~\cite{qwen3speed}} \\
BF16 (baseline) & 47.10\,/\,525.80 & 1.00\,/\,1.00 & 28{,}402\,MB \\
FP8 (W8A8) & 97.11\,/\,880.72 & 2.06\,/\,1.68 & 16{,}012\,MB \\
AWQ-INT4 (W4A16) & 96.49\,/\,744.74 & 2.05\,/\,1.42 & 9{,}962\,MB \\
\midrule
\multicolumn{4}{@{}l}{\textit{vLLM 0.11.0 and SGLang 0.5.4.post2 at default settings, one H100 SXM, gpt-oss-20b,}} \\
\multicolumn{4}{@{}l}{\textit{1000 ShareGPT prompts; total tokens/s}~\cite{gpustackgptoss}} \\
MXFP4, vLLM & 16{,}493.75 & n.r. & 12.8\,GiB$^\star$ \\
MXFP4, SGLang & 5{,}626.99 & n.r. & 12.8\,GiB$^\star$ \\
\bottomrule
\end{tabular}
\end{table}

Table~\ref{tab:perf} has every in-band timing we found: four engine--GPU groups, each with its own model and metric, so speedups compare within a group only.
A memory-versus-compute (roofline) view explains all of them.
At batch size 1, each generated token reads every weight once, so decode time is set by bytes moved, and 4-bit weights move a quarter of the bytes of BF16.
That is why every INT4 recipe is $1.35$--$2.12\times$ faster at batch size 1; the gain falls short of $4\times$ because dequantization, attention, and KV-cache reads do not shrink.
At batch size 256, one read of the weights serves 256 sequences, the matmuls become compute-bound, and W4A16 still computes in 16 bits, so Gemma~3 12B gains only $1.02\times$.
Long prompts are compute-bound for the same reason: the A100 INT4 gain falls from $2.12\times$ to $1.64\times$ at a 30{,}720-token prompt, and on the H20, whose metric counts prompt tokens, FP8 pulls ahead of AWQ ($1.68\times$ versus $1.42\times$) because Hopper's FP8 tensor cores run at twice the BF16 rate~\cite{nvidiah100}.
The rule predicts that FP8 keeps a speedup at large batch where INT4 loses it, but no in-band source tests FP8 above batch size 1.

gpt-oss-20b is a different case: a mixture-of-experts model with 20.9B total and 3.6B active parameters, post-trained with MXFP4 expert weights~\cite{openai2025gptoss}.
Its only in-band timing compares engines, not recipes: at default settings, vLLM delivers $2.93\times$ the total throughput of SGLang on the same prompts, and the same report raises vLLM to 19{,}128.79 tokens/s with asynchronous scheduling~\cite{gpustackgptoss}.
No BF16 timing exists, although MXFP4 values are exactly representable in BF16 at realistic block scales, so a BF16 copy would have the same accuracy.
For AMD, published FP8 tables stop at Llama~3.1 70B and 405B~\cite{rocmblog2025}; Intel's Gaudi~3 data uses Optimum-Habana~\cite{intelgaudi2026}, and Lee et al.\ measure FP8 on Gaudi~2 without vLLM or SGLang~\cite{lee2025gaudi}.

\section{Accuracy: most deltas are within noise}
\label{sec:acc}

\begin{table}[!tp]
\caption{Accuracy against BF16. BF16 rows give the source's score; other rows give the change in points (negative is worse). Each group has its own BF16 baseline because harness settings differ by source; compare deltas within a group. None of these scores comes from a timing run in Table~\ref{tab:perf}. $^\ddagger$HumanEval pass@2 on 64 samples. $^\S$MMLU philosophy subset (311 questions) only; the AWQ checkpoint was calibrated on 30 of its samples.}
\label{tab:acc}
\begin{tabular}{@{}lrrrr@{}}
\toprule
Recipe & MMLU & GSM8K & HumanEval & C-Eval \\
\midrule
\multicolumn{5}{@{}l}{\textit{Qwen3-14B, AngelSlim, lm-eval, engine n.r.}~\cite{angelslim2025}} \\
BF16 & 78.90 & 88.40 & 55.49 & 83.06 \\
FP8-static & $-0.33$ & $+1.06$ & $+1.83$ & $-0.44$ \\
FP8-dynamic & $+0.02$ & $-0.08$ & $-3.05$ & $-0.82$ \\
INT8-dynamic (W8A8) & $-0.77$ & $-2.12$ & $+0.61$ & $-1.19$ \\
GPTQ-INT4 (W4A16) & $-0.88$ & $-1.06$ & $+2.44$ & $-2.01$ \\
AWQ-INT4 (W4A16) & $-1.22$ & $-4.17$ & $+6.10$ & $-1.04$ \\
\midrule
\multicolumn{5}{@{}l}{\textit{Qwen3-14B, Red Hat AI, lm-eval on vLLM}~\cite{rhqwen3fp8,rhqwen3w4a16}} \\
BF16 & 76.81 & 67.63 & n.r. & n.r. \\
FP8-dynamic & $-0.34$ & $-1.44$ & n.r. & n.r. \\
GPTQ W4A16 & $-0.91$ & $+3.94$ & n.r. & n.r. \\
\midrule
\multicolumn{5}{@{}l}{\textit{Qwen3-14B, Red Hat AI, lm-eval on vLLM}~\cite{rhqwen3nvfp4}} \\
BF16 & 78.86 & 88.70 & 90.74$^\ddagger$ & n.r. \\
NVFP4 (W4A4) & $-1.63$ & $-0.45$ & $-0.87^\ddagger$ & n.r. \\
\midrule
\multicolumn{5}{@{}l}{\textit{Qwen2.5-14B-Instruct, Red Hat AI, lm-eval on vLLM}~\cite{rhqwen25w8a8}} \\
BF16 & 79.87 & 83.55 & n.r. & n.r. \\
INT8 W8A8 & $-0.12$ & $+1.97$ & n.r. & n.r. \\
\midrule
\multicolumn{5}{@{}l}{\textit{Gemma~3 12B IT, Red Hat AI, lm-eval on vLLM}~\cite{rhgemmafp8,rhgemmaw8a8}} \\
BF16 & 72.15 & 88.10 & n.r. & n.r. \\
FP8-dynamic & $-0.35$ & $-0.08$ & n.r. & n.r. \\
INT8 W8A8 & $-0.50$ & $-0.38$ & n.r. & n.r. \\
\midrule
\multicolumn{5}{@{}l}{\textit{Gemma~3 12B IT, PyTorch, lm-eval on Transformers}~\cite{gemma3awqcard}} \\
BF16 & 79.10$^\S$ & n.r. & n.r. & n.r. \\
TorchAO INT4 (W4A16) & $-3.54^\S$ & n.r. & n.r. & n.r. \\
AWQ-INT4 (W4A16) & $-2.25^\S$ & n.r. & n.r. & n.r. \\
\midrule
\multicolumn{5}{@{}l}{\textit{Phi-4, Red Hat AI, lm-eval on vLLM}~\cite{rhphi4fp8,rhphi4w8a8,rhphi4w4a16}} \\
BF16 & 80.30 & 90.07 & n.r. & n.r. \\
FP8-dynamic & $0.00$ & $+0.60$ & n.r. & n.r. \\
INT8 W8A8 & $+0.09$ & $+0.23$ & n.r. & n.r. \\
GPTQ W4A16 & $-0.43$ & $-0.38$ & n.r. & n.r. \\
\midrule
\multicolumn{5}{@{}l}{\textit{Mistral NeMo Instruct 2407, Red Hat AI, lm-eval on vLLM}~\cite{rhnemofp8,rhnemow4a16}} \\
BF16 & 68.35 & 74.45 & n.r. & n.r. \\
FP8-static & $+0.15$ & $-1.44$ & n.r. & n.r. \\
GPTQ W4A16 & $-1.43$ & $-4.02$ & n.r. & n.r. \\
\bottomrule
\end{tabular}
\end{table}

Table~\ref{tab:acc} covers five of the seven models; we found no quantized accuracy table for Ministral~3 14B and none for INT4 on Qwen2.5-14B-Instruct.
No source reports a standard error, so we supply a yardstick.
For a score $p$ on $n$ questions, one standard error of a single run is $\sqrt{p(1-p)/n}$: about 3.9 points on HumanEval ($n=164$)~\cite{chen2021humaneval}, 0.8--1.3 on GSM8K ($n=1{,}319$)~\cite{cobbe2021gsm8k}, 0.34--0.39 on full MMLU ($n=14{,}042$)~\cite{hendrycks2021mmlu}, and 2.3 on MMLU philosophy ($n=311$), at the BF16 scores in Table~\ref{tab:acc}.
The error of a delta also depends on how many answers flip between runs~\cite{dutta2024accuracy}, which no source reports.

Against this yardstick, the HumanEval deltas, from $-3.05$ to $+6.10$, are $-5$ to $+10$ problems and cannot separate recipes.
The GSM8K losses of 4.17 (AWQ on Qwen3-14B) and 4.02 (GPTQ on Mistral NeMo) are more than three single-run standard errors and are the clearest losses in the table.
On full MMLU, every 8-bit delta is at most 0.77 points, about two standard errors, and INT4 costs 0.43 to 1.43 points; NVFP4 costs 1.63.
The Gemma INT4 rows are both small-sample and contaminated, since the AWQ checkpoint was calibrated on the subject it is scored on.
Harness settings matter as much as recipes: the two Qwen3-14B groups disagree on GSM8K for the same W4A16 idea ($-1.06$ versus $+3.94$), and their BF16 baselines differ by 20.77 points.
gpt-oss-20b has no delta; its card reports 85.3 on MMLU at high reasoning effort, a protocol not comparable with the rows above~\cite{openai2025gptoss}.

\section{What the evidence supports}
\label{sec:supports}

\begin{table}[!tp]
\caption{What the public record supports for 9B--20B models on vLLM and SGLang, and how firmly. Confidence reflects the number of independent sources, models, and GPUs behind each claim.}
\label{tab:supports}
\begin{tabular}{@{}>{\raggedright\arraybackslash}p{1.25in}>{\raggedright\arraybackslash}p{2.35in}>{\raggedright\arraybackslash}p{1.55in}@{}}
\toprule
Question & What the record shows & Confidence \\
\midrule
Speed at batch size 1 & W4A16 is $1.35$--$2.12\times$ faster than BF16; FP8 ties AWQ on the H20 & Moderate: 3 sources, 3 GPUs, 2024--25 engines \\
\addlinespace
Speed at large batch & W4A16 gains nothing at batch 256; FP8 untested & Low: 1 model, 1 source \\
\addlinespace
Speed on long prompts & FP8 beats AWQ ($1.68\times$ vs $1.42\times$) & Low: 1 model, 1 GPU (H20) \\
\addlinespace
8-bit accuracy & Within 0.77 MMLU points of BF16 & Moderate--high: 5 models, 2 publishers \\
\addlinespace
INT4 accuracy & 0.43--1.43 MMLU points; up to about 4 GSM8K points & Moderate: 3 models, 2 publishers \\
\addlinespace
gpt-oss-20b engine & vLLM $2.93\times$ SGLang at defaults & Low: 1 source, untuned \\
\addlinespace
AMD, Gaudi speed; FP4 speedup & No in-band measurement & None \\
\bottomrule
\end{tabular}
\end{table}

Table~\ref{tab:supports} grades each claim.
For latency-bound serving at small batch, W4A16 cuts per-token latency by a quarter to a half and costs about one MMLU point; check GSM8K, which loses up to four.
For throughput-bound serving at large batch, INT4 is a memory choice, taking a 14B model from about 28\,GB to about 10\,GB, and FP8 is the recipe the roofline favors, but no public in-band timing confirms it.
On Ampere, FP8 checkpoints run only as weight-only, so the roofline predicts gains like GPTQ-Int8's $1.52\times$, not INT4's.

\section{A reporting checklist}
\label{sec:checklist}

Several gaps above trace to details a source omitted rather than to experiments nobody ran.
A quantized-serving claim answers the practitioner's question if it reports:
(1)~one checkpoint ID and revision, used for both timing and accuracy;
(2)~engine, version, and every non-default flag;
(3)~GPU model and count;
(4)~input and output lengths, batch size or request rate, and whether throughput counts prompt tokens;
(5)~a BF16 baseline timed on the same stack;
(6)~harness, version, few-shot count, chat template, and reasoning mode, with $n$ and standard errors;
(7)~the calibration data, confirmed disjoint from the evaluation sets.
No source in this letter meets all seven: none reports standard errors, and only the Gemma card times the checkpoint it scores.

\section{Limitations}

The rows in Table~\ref{tab:perf} come from different experiments and are not one benchmark.
Our standard errors assume independent questions and a single run.
The Qwen FP8 checkpoint used for timing is quantized in blocks of 128~\cite{qwen3fp8card}, but its activation scaling is not stated, so it cannot be matched to the static or dynamic accuracy rows.
All web sources are vendor or project pages that can change; they were last checked on 5 October 2026.

\bibliographystyle{ACM-Reference-Format}
\bibliography{refs}

\end{document}